\documentclass[letterpaper]{article}
\usepackage{uai2020}
\usepackage{microtype}
\usepackage{graphicx}
\usepackage{float}

\usepackage{multicol}
\usepackage{lipsum}
\usepackage{subfigure}
\usepackage{times}
\usepackage[round]{natbib}

\usepackage{soul}
\usepackage{url}
\usepackage[hidelinks]{hyperref}
\usepackage[utf8]{inputenc}
\usepackage[small]{caption}
\usepackage{graphicx}
\usepackage{amsmath}
\usepackage{amsfonts}
\usepackage{booktabs}
\DeclareMathOperator*{\argmax}{arg\,max}

\usepackage[margin=1in]{geometry}

\usepackage{times}

\title{Batch norm with entropic regularization turns deterministic autoencoders into generative models}

\author{Amur Ghose \and Abdullah Rashwan \and Pascal Poupart \\
University of Waterloo and Vector Institute, Ontario, Canada \\
\{a3ghose,arashwan,ppoupart\}@uwaterloo.ca} % LEAVE BLANK FOR ORIGINAL SUBMISSION.
\begin{document}

\maketitle

\begin{abstract}
The variational autoencoder is a well defined deep generative model that utilizes an encoder-decoder framework where an encoding neural network outputs a non-deterministic code for reconstructing an input. The encoder achieves this by sampling from a distribution for every input, instead of outputting a deterministic code per input. The great advantage of this process is that it allows the use of the network as a generative model for sampling from the data distribution beyond provided samples for training. We show in this work that utilizing batch normalization as a source for non-determinism suffices to turn deterministic autoencoders into generative models on par with variational ones, so long as we add a suitable entropic regularization to the training objective.
\end{abstract}

\section{INTRODUCTION}

Modeling data with neural networks is often broken into the broad classes of discrimination and generation. We consider generation, which can be independent of related goals like density estimation, as the task of generating unseen samples from a data distribution, specifically by neural networks, or simply \textbf{deep generative models}.

The variational autoencoder \citep{kingma2013auto} (VAE) is a well-known subclass of deep generative models, in which we have two distinct networks - a decoder and encoder.  To generate data with the decoder, a sampling step is introduced between the encoder and decoder.  This sampling step complicates the optimization of autoencoders. Since it is not possible to differentiate through sampling, the reparametrization trick is often used.  The sampling distribution has to be optimized to approximate a canonical distribution such as a Gaussian.  The log-likelihood objective is also approximated.  Hence, it would be desirable to avoid the sampling step. 

To that effect, \cite{ghosh2019variational} proposed regularized autoencoders (RAEs)  where sampling is replaced by some regularization, since stochasticity introduced by sampling can be seen as a form of regularization.  By avoiding  sampling, a deterministic autoencoder can be optimized more simply. However, they introduce multiple candidate regularizers, and picking the best one is not straightforward. Density estimation also becomes an additional task as \cite{ghosh2019variational} fit a density to the empirical latent codes after the autoencoder has been optimized. 

In this work, we introduce a batch normalization step between the encoder and decoder and add a entropic regularizer on the batch norm layer.  Batchnorm fixes some moments (mean and variance) of the empirical code distribution while the entropic regularizer maximizes the entropy of the empirical code distribution.  Maximizing the entropy of a distribution with certain fixed moments induces Gibbs distributions of certain families (i.e., normal distribution for fixed mean and variance).  Hence, we naturally obtain a distribution that we can sample from to obtain codes that can be decoded into realistic data.  The introduction of a batchnorm step with entropic regularization does not complicate the optimization of the autoencoder which remains deterministic. Neither step is sufficient in isolation and requires the other, and we compare what happens when the entropic regularizer is absent. Our work parallels RAEs in determinism and regularization, though we differ in choice of regularizer and motivation, as well as ease of isotropic sampling.

The paper is organized as follows.  In Section~\ref{sec:background}, we review background about variational autoencoders and batch normalization.  In Section~\ref{sec:eae}, we propose entropic autoencoders (EAEs) with batch normalization as a new deterministic generative model.  Section~\ref{sec:maxent} discusses the maximum entropy principle and how it promotes certain distributions over latent codes even without explicit entropic regularization.  Section~\ref{sec:experiments} demonstrates the generative performance of EAEs on three benchmark datasets (CELEBA, CIFAR-10 and MNIST).  EAEs outperform previous deterministic and variational autoencoders in terms of FID scores.  Section~\ref{sec:conclusion} concludes the paper with suggestions for future work.

\section{VARIATIONAL AUTOENCODER}
\label{sec:background}

The variational autoencoder \citep{kingma2013auto} (VAE) consists of a decoder followed by an encoder. The term autoencoder \citep{ng2011sparse} is in general applied to any model that is trained to reconstruct its inputs. For a normal autoencoder, representing the decoder and encoder as $\mathcal{D}, \mathcal{E}$ respectively, for every input $x_i$ we seek:
$$ \mathcal{E}(x_i) = z_i, \mathcal{D}(z_i) = \hat{x_i} \approx x_i $$
Such a model is usually trained by minimizing $||\hat{x_i} - x_i||^2$ over all $x_i$ in training set. In a variational autoencoder, there is no fixed codeword $z_i$ for a $x_i$. Instead, we have
$$ z_i = \mathcal{E}(x_i) \sim \mathcal{N} (\mathcal{E}_{\mu}(x_i), \mathcal{E}_{\sigma^2} (x_i)) $$
The encoder network calculates means and variances via $\mathcal{E}_{\mu}, \mathcal{E}_{\sigma^2}$ layers for every data instance, from which a code is sampled. The loss function is of the form:
$$||\mathcal{D}(z_i) - x_i||^2 + \beta D_{KL} ( \mathcal{N} (\mathcal{E}_{\mu}(x_i), \mathcal{E}_{\sigma^2} (x_i)) || \mathcal{N} (0,I) )  $$
where $D_{KL}$ denotes the Kullback-Leibler divergence and $z_i$ denotes the sample from the distribution over codes. Upon minimizing the oss function over $x_i \in$ a training set, we can generate samples as : generate $z_i \sim \mathcal{N} (0,I)$, and output $\mathcal{D}(z_i)$. The KL term makes the implicitly learnt distribution of the encoder close to a spherical Gaussian. Usually, $z_i$ is of a smaller dimensionality than $x_i$.

\subsection{VARIATIONS ON VARIATIONAL AUTOENCODERS}

In practice, the above objective is not easy to optimize. The original VAE formulation did not involve $\beta$, and simply set it to $1$. Later, it was discovered that this parameter helps training the VAE correctly, giving rise to a class of architectures termed $\beta$-VAE. \citep{higgins2017beta}

The primary problem with the VAE lies in the training objective. We seek to minimize KL divergence for every instance $x_i$, which is often too strong. The result is termed \textbf{posterior collapse} \citep{he2019lagging} where every $x_i$ generates $\mathcal{E}_{\mu}(x_i) \approx 0, \mathcal{E}_{\sigma^2}(x_i) \approx 1$. Here, the latent variable $z_i$ begins to relate less and less to $x_i$, because neither $\mu,\sigma^2$ depend on it. Attempts to fix this \citep{kim2018semi} involve analyzing the mutual information between $z_i,x_i$ pairs, resulting in architectures like InfoVAE \citep{zhao2017infovae}, along with others such as $\delta$-VAE \citep{razavi2019preventing}. Posterior collapse is notable when the decoder is especially `powerful', i.e. has great representational power. Practically, this manifests in the decoder's depth being increased, more deconvolutional channels, etc.

One VAE variation includes creating a deterministic architecture that minimizes an optimal transport based Wasserstein loss between the empirical data distribution and decoded images from aggregate posterior. Such models \citep{tolstikhin2017wasserstein} work with the aggregate posterior instead of outputting a distribution per sample, by optimizing either the Maximum Mean Discrepancy (MMD) metric with a Gaussian kernel \citep{gretton2012kernel}, or using a GAN to minimize this optimal transport loss via Kantorovich-Rubinstein duality. The GAN variant outperforms using MMD, and WAE techniques are usually considered as WAE-GAN for achieving state-of-the-art results.

\subsection{BATCH NORMALIZATION}

Normalization is often known in statistics as the procedure of subtracting the mean of a dataset and dividing by the standard deviation. This sets the sample mean to zero and variance to one. In neural networks, normalization for a minibatch \citep{ioffe2015batch} has become ubiquitous since its introduction and is now a key part of training all forms of deep generative models \citep{ioffe2017batch}. Given a minibatch of inputs $x_i$ of dimensions $n$ with $\mu_{ij}, \sigma_{ij}$ as its mean, standard deviation at index $j$ respectively, we will call \textbf{BN} as the operation that satisfies:
\begin{equation} \label{bn}
[\textrm{{BN}}(x_i)]_j = \frac{x_{ij} - \mu_{ij}}{\sigma_{ij}} 
\end{equation}
Note that in practice, a batch normalization layer in a neural network computes a function of form $A \circ B$ with $A$ as an affine function, and $B$ as \textbf{BN}. This is done during training time using the empirical average of the minibatch, and at test time using the overall averages. Many variations on this technique such as L1 normalization, instance normalization, online adaptations, etc. exist \citep{wu2018l1,zhang2019fixup,chiley2019online,ulyanov2016instance,ba2016layer,hoffer2018norm}. The mechanism by which this helps optimization was initially termed as ``internal covariate shift", but later works challenge this perception \citep{santurkar2018does,yang2019mean} and show it may have harmful effects \citep{galloway2019batch}.

%\section{Wasserstein Autoencoders}

%One of the variations on VAEs includes creating a deterministic architecture that minimizes an optimal transport based Wasserstein loss between the latent space and the prior. Such an architecture \citep{tolstikhin2017wasserstein} benefits from not creating a distribution for every sample. Instead, the entire posterior space is made to be close to a Gaussian distribution. This can be done by optimizing either the Maximum Mean Discrepancy (MMD) metric with a Gaussian kernel \citep{gretton2012kernel}, or using a GAN to discriminate between samples generated from the prior (Gaussian) or the empirical posterior, however, in practice, the GAN offers too large a performance increase to pass up, causing WAE techniques to be usually considered as WAE-GAN.

\section{OUR CONTRIBUTIONS - THE ENTROPIC AUTOENCODER}
\label{sec:eae}

Instead of outputting a distribution as VAEs do, we seek an approach that turns deterministic autoencoders into generative models on par with VAEs. Now, if we had a guarantee that, for a regular autoencoder that merely seeks to minimize reconstruction error, the distribution of all $z_i$'s approached a spherical Gaussian, we could carry out generation just as in the VAE model. We do the following : we simply append a batch normalization step (BN as above, i.e. no affine shift) to the end of the encoder, and minimize the objective:
\begin{equation}
     ||\hat{x_i} - x_i ||^2 - \beta H (z_i), \hat{x_i} = \mathcal{D}(z_i), z_i = \mathcal{E}(x_i)
     \label{eq:objective}
\end{equation}
where $H$ represents the entropy function and is taken over a minibatch of the $z_i$. We recall and use the following property : let $X$ be a random variable obeying $E[X] = 0, E[X^2] = 1$. Then, the maximum value of $H(X)$ is obtained iff $X \sim \mathcal{N} (0,1)$. We later show that even when no entropic regularizer is applied, batch norm alone can yield passable samples when a Gaussian is used for generation purposes. However, for good performance, the entropic regularizer is necessary.

\subsection{EQUIVALENCE TO KL DIVERGENCE MINIMIZATION}

Our method of maximizing entropy minimizes the KL divergence by a backdoor. Generally, minibatches are too small to construct a meaningful sample distribution that can be compared - in $D_{KL}$ - to the sought spherical normal distribution without other constraints. However, suppose that we have the following problem with $X$ being a random variable with some constraint functions $C_k$ e.g. on its moments:

$$ \max H(X), E[C_k(X)] = c_k, k = 1,2,\dots $$

In particular let the two constraints be $E[X] = 0, E[X^2] = 1$ as above. Consider a `proposal' distribution $Q$ that satisfies $E_Q[X] = 0, E_Q[X^2] = 1$ and also a maximum entropy distribution $P$ that is the solution to the optimization problem above. The cross entropy of $P$ with respect to $Q$ is

$$ E_Q [- \log P(X) ] $$

In our case, $P$ is a Gaussian and $-\log P(X)$ is a term of the form $aX^2 + bX +c$. In expectation of this w.r.t. $Q$, $E_Q[X], E_Q[X^2]$ are already fixed. Thus for all proposal distributions $Q$, cross entropy of $P$ w.r.t. $Q$ - written as $H(Q,P)$ obeys

$$ H(Q,P) = H(Q) + D_{KL} (Q||P)$$

Pushing up $H(Q)$ thus directly reduces the KL divergence to $P$, as the left hand side is a constant. Over a minibatch, every proposal $Q$ identically satisfies the two moment conditions due to normalization. Unlike KL divergence, involving estimating and integrating a conditional probability (both rapidly intractable in higher dimensions) entropy estimation is easier, involves no conditional probabilities, and forms the bedrock of estimating quantities derived from entropy such as MI. Due to interest in the Information Bottleneck method \citep{tishby2000information} which requires entropy estimation of hidden layers, we already have a nonparametric entropy estimator of choice - the Kozachenko Leonenko estimator \citep{kozachenko1987sample}, which also incurs low computational load and has already been used for neural networks. This principle of ``cutting the middleman" builds on the fact that MI based methods for neural networks often use Kraskov-like estimators \citep{kraskov2004estimating}, a family of estimators that break the MI term into H terms which are estimated by the Kozachenko-Leonenko estimator. Instead, we directly work with the entropy.

\subsection{A GENERIC NOTE ON THE KOZACHENKO-LEONENKO ESTIMATOR}

The Kozachenko-Leonenko estimator \citep{kozachenko1987sample} operates as follows. Let $N \geq 1$ and $X_1,\dots,X_{N+1}$ be i.i.d. samples from an unknown distribution $Q$. Let each $X_i \in \mathbb{R}^d$.

For each $X_i$, define $R_i = \min ||X_i-X_j||_2, j \neq i$ and $Y_i = N (R_i)^d$. Let $B_d$ be the volume of the unit ball in $\mathbb{R}^d$ and $\gamma$ the Euler-mascheroni constant $\approx 0.577$. The Kozachenko Leonenko estimator works as follows:
$$ H(Q) \approx \frac{1}{N+1} \sum_{i=1}^{N+1} \log Y_i + \log B_d + \gamma $$
Intuitively, having a high distance to the nearest training example for each example pushes up the entropy via the $Y_i$ term. Such ``repulsion"-like nearest neighbour techniques have been employed elsewhere for likelihood-free techniques such as implicit maximum likelihood estimation \citep{li2019diverse,li2018implicit}. In general, the estimator is biased with known asymptotic orders \citep{delattre2017kozachenko} - however, when the bias stays relatively constant through training, optimization is unaffected. The complexity of the estimator when utilizing nearest neighbours per minibatch is quadratic in the size of the batch, which is reasonable for small batches.

\subsection{GENERALIZATION TO ANY GIBBS DISTRIBUTION}

A distribution that has the maximum entropy under constraints $C_k$ as above is called the \textbf{Gibbs distribution} of the respective constraint set. When this distribution exists, we have the result that there exist Lagrange multipliers $\lambda_k$, such that if the maximum entropy distribution is $P$, $\log P (X)$ is of the form $\sum \lambda_k C_k$. For any candidate distribution $Q$, $E_Q[C_k(X)]$ is determined solely from the constraints, and thus the cross-entropy $E_Q[-\log P(X)]$ is also determined. Our technique of pushing up the entropy to reduce KL holds under this generalization. For instance, pushing up the entropy for $L_1$ normalization layers corresponds to inducing a Laplace distribution.

\subsection{PARALLELS WITH THE CONSTANT VARIANCE VAE}

One variation on VAEs is the constant variance VAE \citep{ghosh2019variational}, where the term $\mathcal{E}_{\sigma^2}$ is constant for every instance $x_i$. Writing Mutual Information as MI, consider transmitting a code via the encoder that maximizes $\textrm{MI}(X,Y)$ where $X$ is the encoder's output and $Y$ the input to the decoder. In the noiseless case, $Y=X$, and we work with $\textrm{MI}(X,X)$.

For a discrete random variable $X$, $\textrm{MI}(X,X) = H(X)$. If noiseless transmission was possible, the mutual information would depend solely on entropy. However, using continuous random variables, our analysis of the constant variance autoencoder would for $\sigma^2 = 0$ yield a MI of $\infty$, between the code emitted by the encoder and received by the decoder. This at first glance appears ill-defined.

However, suppose that we are in the test conditions i.e. the batch norm is using a fixed mean and variance and independent of minibatch. Now, if the decoder receives $Y$, $\textrm{MI}(X,Y) = H(X) - H(X|Y)$. Since $(X|Y)$ is a Dirac distribution, it pushes the mutual information to $\infty$. If we ignore the infinite mutual information introduced by the deterministic mapping just as in the definition of differential entropy, the only term remaining is $H(X)$, maximizing which becomes equivalent to maximizing MI. We propose our model as the zero-variance limit of present constant variance VAE architectures, especially when batch size is large enough to allow accurate estimations of mean and variance.

\subsection{COMPARISON TO PRIOR DETERMINISTIC AUTOENCODERS }

Our work is not the first to use a deterministic autoencoder as a generative one. Prior attempts in this regard such as regularized autoencoders (RAEs) \citep{ghosh2019variational} share the similarities of being deterministic and regularized autoencoders, but do not leverage batch normalization. Rather, these methods rely on taking the constant variance autoencoder, and imposing a regularization term on the architecture. This does not maintain the KL property that we show arises via entropy maximization, rather, it forms a latent space that has to be estimated such as via a Gaussian mixture model (GMM) on top of the regularization. The Gaussian latent space is thus lost, and has to be estimated post-training. In contrast to the varying regularization choices of RAEs, our method uses the specific Max Entropy regularizer forcing a particular latent structure.  Compared to the prior Wasserstein autoencoder (WAE) \citep{tolstikhin2017wasserstein}, RAEs achieve better empirical results, however we further improve on these results while keeping the ability to sample from the prior i.e. isotropic Gaussians. As such, we combine the ability of WAE-like sampling with performance superior to RAEs, delivering the best of both worlds. This comparison excludes the much larger bigWAE models \citep{tolstikhin2017wasserstein} which utilize ResNet encoder-decoder pairs.

In general, for all VAE and RAE-like models, the KL/Optimal Transport/Regularization terms compete against reconstruction loss and having perfect Gaussian latents is not always feasible, hence, EAEs, like RAEs, benefit from post-density estimation and GMM fitting. The primary advantage they attain is not \textbf{requiring} such steps, and performing at a solid baseline without it.

\section{THE MAXIMUM ENTROPY PRINCIPLE AND REGULARIZER-FREE LATENTS}
\label{sec:maxent}

We now turn to a general framework that motivates our architecture and adds context. Given the possibility of choosing a distribution $Q \in \mathcal{D}$ that fits some given dataset $\mathcal{X}$ provided, what objective should we choose? One choice is to pick:
$$ Q = \argmax_{Q \in \mathcal{D}} E_{\bar{\mathcal{X}}}[LL_Q(X)] $$
where $LL_Q(X)$ denotes the log likelihood of an instance $X$ and $E_{\bar{\mathcal{X}}}$ indicates that the expectation is taken with the empirical distribution  $\bar{\mathcal{X}}$ from $\mathcal{X}$, i.e. every point $X$ is assigned a probability $\frac{1}{| \mathcal{X}|}$. An alternative is to pick:
\begin{align}
Q = & \argmax_{Q \in \mathcal{D}} H(Q) \\
& \mbox{subject to } T_i(Q) = T_i(\mathcal{X})
\end{align}
Where $H$ is the entropy of $Q$, and $T_i(Q)$ are summary statistics of $Q$ that match the summary statistics over the dataset. For instance, if all we know is the mean and variance of $\mathcal{X}$, the distribution $Q$ with maximum entropy that has the same mean and variance is Gaussian. This so-called maximum entropy principle \citep{bashkirov2004maximum} has been used in reinforcement learning \citep{ziebart2008maximum}, natural language processing \citep{berger1996maximum}, normalizing flows \citep{loaiza2017maximum}, and computer vision \citep{skilling1984maximum} successfully. Maximum entropy is in terms of optimization the convex dual problem of maximum likelihood, and takes a different route of attacking the same objective.

\subsection{THE MAXENT PRINCIPLE APPLIED TO DETERMINISTIC AUTOENCODERS}

Now, consider the propagation of an input through an autoencoder. The autoencoder may be represented as:
$$ X \approx \mathcal{D}(\mathcal{E}(X)) $$
where $\mathcal{D},\mathcal{E}$ respectively represent the decoder and encoder halves. Observe that if we add a BatchNorm of the form $A \circ B$ with $A$ as an affine shift, $B$ as \textbf{BN} (as defined in Equation \ref{bn}) to $\mathcal{E}$ - the encoder - we try to find a distribution $Z$ after $B$ and before $A$, such that:
\begin{itemize}
\item $E[Z] = 0, E[Z^2] = 1$
\item $B \circ \mathcal{E}(X) \sim Z$, $A \circ \mathcal{D}(Z) \sim X$
\end{itemize}
Observe that there are two conditions that do not depend on $\mathcal{E},\mathcal{D}$: $E[Z] = 0, E[Z^2] = 1$. Consider two different optimization problems:

\begin{itemize}
\item $O$, which asks to find the max entropy distribution $Q$, i.e., with max $H(Q)$ over $Z$ satisfying $E_Q[Z] = 0, E_Q[Z^2] = 1$.
\item $O'$, which asks to find $\mathcal{D},\mathcal{E},A$ and a distribution $Q'$ over $Z$ such that we maximize $H(Q')$, with $E_{Q'}[Z] = 0, E_{Q'}[Z^2] = 1, B \circ \mathcal{E}(X) \sim Z, A \circ \mathcal{D}(Z) \sim X, Z \sim Q'$.
\end{itemize}

Since $O$ has fewer constraints, $H(Q) \geq H(Q')$. Furthermore, $H(Q)$ is known to be maximal iff $Q$ is an isotropic Gaussian over $Z$. What happens as the capacity of $\mathcal{D},\mathcal{E}$ rises to the point of possibly representing anything (e.g., by increasing depth)? The constraints $B \circ \mathcal{E}(X) \sim Z, A \circ \mathcal{D}(Z) \sim X, Z \sim Q'$ effectively vanish, since the functional ability to deform $Z$ becomes arbitrarily high.  We can take the solution of $O$, plug it into $O'$, and find $\mathcal{E},\mathcal{D},A$ that (almost) meet the constraints of $B \circ \mathcal{E}(X) \sim Z, A \circ \mathcal{D}(Z) \sim X, Z \sim Q'$. If the algorithm chooses the max entropy solution, the solution of $O'$ - the actual distribution after the BatchNorm layer - approaches the maxent distribution, an isotropic Gaussian, when the last three constraints in $O$ affect the solution less.

\subsection{NATURAL EMERGENCE OF GAUSSIAN LATENTS IN DEEP NARROWLY BOTTLENECKED AUTOENCODERS}

We make an interesting prediction: if we increase the depths of $\mathcal{E},\mathcal{D}$ and constrain $\mathcal{E}$ to output a code $Z$ obeying $E[Z] = 0, E[Z^2] = 1$, the distribution of $Z$ should - even without an entropic regularizer - tend to go to a spherical Gaussian as depth increases relative to the bottleneck. In practical terms, this will manifest in less regularization being required at higher depths or narrower bottlenecks. This phenomenon also occurs in posterior collapse for VAEs and we should verify that our latent space stays meaningful under such conditions.

Under the information bottleneck principle, for a neural network with output $Y$ from input $X$, we seek a hidden layer representation for $Z$ that maximizes $\textrm{MI}(Z,Y)$ while lowering $\textrm{MI}(X,Z)$. For an autoencoder, $Y \approx X$. Since $Z$ is fully determined from $X$ in a deterministic autoencoder, increasing $H(Z)$ increases $\textrm{MI}(Z,X)$ if we ignore the $\infty$ term that arises due to $H(Y|X)$ as $Y$ approaches a deterministic function of $X$ as before in our CV-VAE discussion. Increasing $H(Z)$ will be justified iff it gives rise to better reconstruction, i.e. making $Z$ more entropic (informative) lowers the reconstruction loss.

Such increases are likelier when $Z$ is of low dimensionality and struggles to summarize $X$. We predict the following: a deep, narrowly bottlenecked autoencoder with a batch normalized code, will, even without regularization, approach spherical Gaussian-like latent spaces. We show this in the datasets of interest, where narrow enough bottlenecks can yield samples even \textbf{without} regularization, a behaviour also anticipated in \citep{ghosh2019variational}.

\section{EMPIRICAL EXPERIMENTS}
\label{sec:experiments}

\subsection{BASELINE ARCHITECTURES WITH ENTROPIC REGULARIZATION}

We begin by generating images based on our architecture on 3 standard datasets, namely MNIST \citep{lecun2010mnist}, CIFAR-10 \citep{krizhevsky2014cifar} and CelebA \citep{liu2018large}. %The architectures we use are exactly as in previous work for deterministic autoencoders, to allow a fair comparison. 
We use convolutional channels of $[128,256,512,1024]$ in the encoder half and deconvolutional channels of $[512,256]$ for MNIST and CIFAR-10 and $[512,256,128]$ for CelebA, starting from a channel size of $1024$ in the decoder half. For kernels we use $4 \times 4$ for CIFAR-10 and MNIST, and $5 \times 5$ for CelebA with strides of $2$ for all layers except the terminal decoder layer. Each layer utilizes a subsequent batchnorm layer and ReLU activations, and the hidden bottleneck layer immediately after the encoder has a batch norm without affine shift. These architectures, preprocessing of datasets, etc. match exactly the previous architectures that we benchmark against \citep{ghosh2019variational, tolstikhin2017wasserstein}.

For optimization, we utilize the ADAM optimizer. The minibatch size is set to $100$, to match \citep{ghosh2019variational} with an entropic regularization based on the Kozachenko Leonenko estimator \citep{kozachenko1987sample}. In general, larger batch sizes yielded better FID scores but harmed speed of optimization. In terms of latent dimensionality, we use $16$ for MNIST, $128$ for CIFAR-10 and $64$ for CelebA. At most $100$ epochs are used for MNIST and CIFAR-10 and at most $70$ for CelebA.

\begin{figure*}[h]
\centering
\includegraphics[width=.3\textwidth]{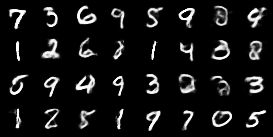}\hfill
\includegraphics[width=.3\textwidth]{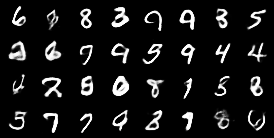}\hfill
\includegraphics[width=.3\textwidth]{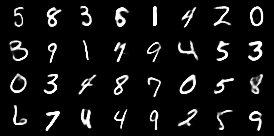}

    \caption{Left: Generated MNIST images with $\beta=0.05$ in Eq.~\ref{eq:objective}.  Middle: Generated MNIST images with $\beta=1.0$ in Eq.~\ref{eq:objective}. Right: Reconstructed MNIST images with $\beta=1.0$ in Eq.~\ref{eq:objective}.}
    \label{MNIST}
\end{figure*}

In Figure \ref{MNIST}, we present qualitative results on the MNIST dataset. We do not report the Frechet Inception Distance (FID) \citep{heusel2017gans}, a commonly used metric for gauging image quality, since it uses the Inception network, which is not calibrated on grayscale handwritten digits. In Figure \ref{MNIST}, we show the quality of the generated images for two different regularization weights $\beta$ in Eq.~\ref{eq:objective} (0.05 and 1.0 respectively) and in the same figure illustrate the quality of reconstructed digits.

\begin{figure}[h]
    \centering
    \includegraphics[keepaspectratio,width=6 cm]{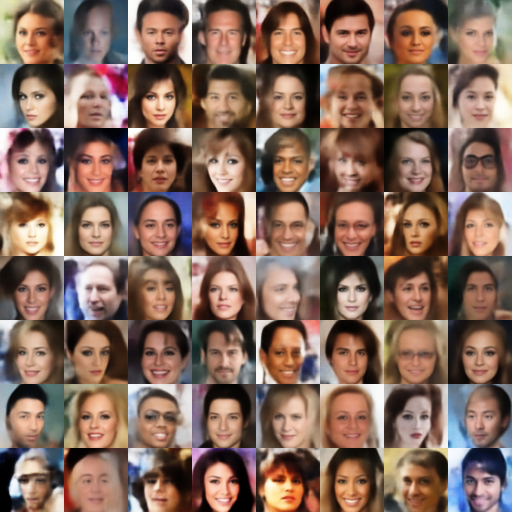}
    \caption{Generated images on CelebA}
    \label{CelebA}
\end{figure}

\begin{figure}[h]
    \centering
    \includegraphics[keepaspectratio,width=3.91 cm]{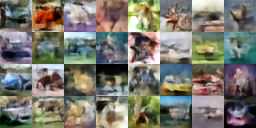}
    \includegraphics[keepaspectratio,width=3.91 cm]{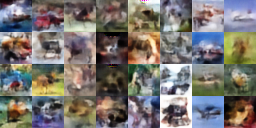}
    
    \vspace{0.1cm}
    
    \includegraphics[keepaspectratio,width=3.91 cm]{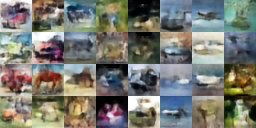}
    \includegraphics[keepaspectratio,width=3.91 cm]{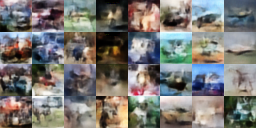}
    \caption{Generated images on CIFAR-10, under four different regularization weights (top left $\beta=0.5$, top right $\beta=0.7$, bottom left $\beta=0.05$, bottom right $\beta=0.07$).}
    \label{CIFAR}
\end{figure}

We move on to qualitative results for CelebA. We present a collage of generated samples in Figure \ref{CelebA}. CIFAR-10 samples are presented in Figure \ref{CIFAR}. We also seek to compare, thoroughly, to the RAE architecture. For this, we present quantitative results in terms of FID scores in Table \ref{FIDscores} (larger version in supplement). We show results when sampling latent codes from an isotropic Gaussian as well as from densities fitted to the empirical distribution of latent codes after the AE has been optimized. We consider isotropic Gaussians and Gaussian mixture models (GMMs). In all cases, we improve on RAE-variant architectures proposed previously \citep{ghosh2019variational}. We refer to our architecture as the \textbf{entropic autoencoder (EAE)}. There is a tradeoff between Gaussian latent spaces and reconstruction loss, and results always improve with ex-post density estimation due to prior-posterior mismatch.

\subsection*{DETAILS ON PREVIOUS TECHNIQUES}

In the consequent tables and figures, VAE/AE have their standard meanings. AE-L2 refers to an autoencoder with only reconstruction loss and L2 regularization, 2SVAE to the Two-Stage VAE as per \citep{dai2019diagnosing}, WAE to the Wasserstein Autoencoder as per \citep{tolstikhin2017wasserstein}, RAE to the Regularized Auto-encoder as per \citep{ghosh2019variational}, with RAE-L2 referring to such with a L2 penalty, RAE-GP to such with a Gradient Penalty, RAE-SN to such with spectral normalization. We use spectral normalization in our EAE models for CelebA, and L2 regularization for CIFAR-10.

\begin{table*}
\centering
\begin{tabular}{l|ll|ll}
\hline
& CIFAR-10 & & CelebA & \\
\hline
Architectures(Isotropic)  & FID  & Reconstruction & FID & Reconstruction  \\
\hline
VAE       & 106.37  & 57.94  & 48.12 &  39.12    \\
CV-VAE       &  94.75  & 37.74   & 48.87 &  40.41  \\
WAE    &  117.44  &  35.97  & 53.67& \textbf{34.81}     \\
2SVAE   &  109.77  & 62.54  & 49.70 &  42.04  \\
EAE & {\bf 85.26(84.53)} & 29.77  & {\bf 44.63} &40.26   \\
\hline
Architectures(GMM)  & FID  & Reconstruction & FID & Reconstruction \\
\hline
RAE   &  76.28  & 29.05   & 44.68 & 40.18     \\
RAE-L2  &  74.16  & 32.24    & 47.97 &  43.52    \\
RAE-GP     &  76.33 & 32.17  &  45.63  &   39.71    \\
RAE-SN    &   75.30 & \textbf{27.61}  & 40.95  &   36.01  \\
AE     &  76.47 & 30.52 & 45.10   &   40.79    \\
AE-L2  &  75.40 & 34.35  & 48.42  &   44.72   \\

EAE &  {\bf 73.12} & 29.77  & {\bf 39.76} &  40.26  \\
\hline
\end{tabular}
\caption{FID scores for relevant VAEs \& VAE-like architectures. Scores within parentheses for EAE denote regularization on a linear map. Isotropic denotes samples drawn from latent spaces of $\mathcal{N} (0,I)$. GMM denotes sampling from a mixture of $10$ Gaussians of full covariance. These evaluations correspond to analogous benchmarking for RAEs \citep{ghosh2019variational}. Larger version in supplement. }
\label{FIDscores}
\end{table*}

\subsection*{QUALITATIVE COMPARISON TO PREVIOUS TECHNIQUES}

\begin{figure*}[!th]
  \centering
  \scalebox{.9}
{\setlength\tabcolsep{3pt}
\begin{sc}
  \small
    \begin{tabular}{ r c c c}
    \toprule
        & Reconstructions & Random Samples & Interpolations \\
        \midrule
      \raisebox{12pt}{GT}
      & \includegraphics[trim={ 0 640 0
        0},clip,width=0.32\linewidth]{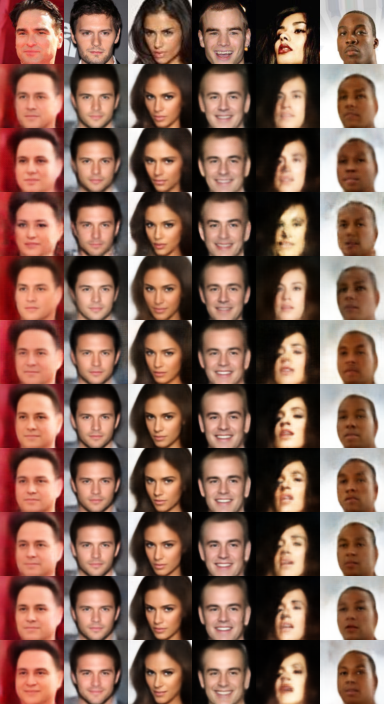}
      \\[-3.5pt]

      \raisebox{12pt}{VAE}
      & \includegraphics[trim={ 0 576 0 64},clip,width=0.32\linewidth]{reconstructionsceleba.png}&\includegraphics[trim={ 0 576 0 0},clip,width=0.32\linewidth]{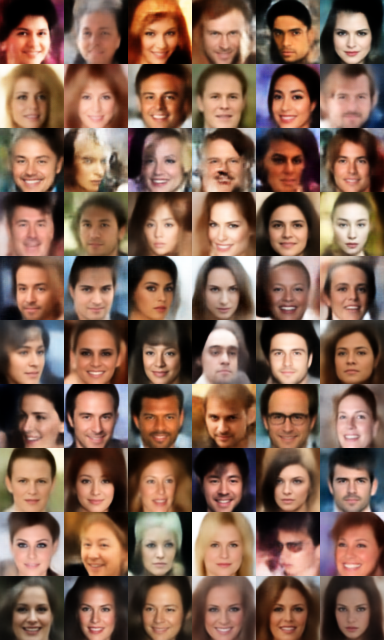} & \includegraphics[trim={ 0 432 0 0},clip,width=0.32\linewidth]{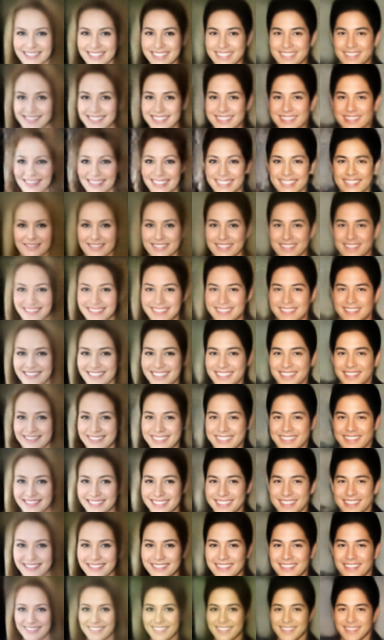} \\[-3.5pt]

      \raisebox{12pt}{CV-VAE}
      & \includegraphics[trim={ 0 512 0 128},clip,width=0.32\linewidth]{reconstructionsceleba.png}&\includegraphics[trim={ 0 512 0 64},clip,width=0.32\linewidth]{image_2020_02_18T14_17_08_899Z.png} & \includegraphics[trim={ 0 384 0 48},clip,width=0.32\linewidth]{interpolbefore.png} \\[-3.5pt]

      \raisebox{12pt}{WAE}
      & \includegraphics[trim={ 0 448 0 192},clip,width=0.32\linewidth]{reconstructionsceleba.png}&\includegraphics[trim={ 0 448 0 128},clip,width=0.32\linewidth]{image_2020_02_18T14_17_08_899Z.png} & \includegraphics[trim={ 0 336 0 96},clip,width=0.32\linewidth]{interpolbefore.png} \\[-3.5pt]

      \raisebox{12pt}{2SVAE}
      & \includegraphics[trim={ 0 384 0 256},clip,width=0.32\linewidth]{reconstructionsceleba.png}&\includegraphics[trim={ 0 384 0 192},clip,width=0.32\linewidth]{image_2020_02_18T14_17_08_899Z.png} & \includegraphics[trim={ 0 288 0 144},clip,width=0.32\linewidth]{interpolbefore.png} \\[-3.5pt]

      \raisebox{12pt}{RAE-GP}
      & \includegraphics[trim={ 0 320 0 320},clip,width=0.32\linewidth]{reconstructionsceleba.png}&\includegraphics[trim={ 0 320 0 256},clip,width=0.32\linewidth]{image_2020_02_18T14_17_08_899Z.png} & \includegraphics[trim={ 0 240 0 192},clip,width=0.32\linewidth]{interpolbefore.png} \\[-3.5pt]

      \raisebox{12pt}{RAE-L2}
      & \includegraphics[trim={ 0 256 0 384},clip,width=0.32\linewidth]{reconstructionsceleba.png}&\includegraphics[trim={ 0 256 0 320},clip,width=0.32\linewidth]{image_2020_02_18T14_17_08_899Z.png} & \includegraphics[trim={ 0 192 0 240},clip,width=0.32\linewidth]{interpolbefore.png} \\[-3.5pt]

      \raisebox{12pt}{RAE-SN}
      & \includegraphics[trim={ 0 192 0 448},clip,width=0.32\linewidth]{reconstructionsceleba.png}&\includegraphics[trim={ 0 192 0 384},clip,width=0.32\linewidth]{image_2020_02_18T14_17_08_899Z.png} & \includegraphics[trim={ 0 144 0 288},clip,width=0.32\linewidth]{interpolbefore.png} \\[-3.5pt]

      \raisebox{12pt}{RAE}
      & \includegraphics[trim={ 0 128 0 512},clip,width=0.32\linewidth]{reconstructionsceleba.png}&\includegraphics[trim={ 0 128 0 448},clip,width=0.32\linewidth]{image_2020_02_18T14_17_08_899Z.png} & \includegraphics[trim={ 0 96 0 336},clip,width=0.32\linewidth]{interpolbefore.png} \\[-3.5pt]

      \raisebox{12pt}{AE}
      & \includegraphics[trim={ 0 64 0 576},clip,width=0.32\linewidth]{reconstructionsceleba.png}&\includegraphics[trim={ 0 64 0 512},clip,width=0.32\linewidth]{image_2020_02_18T14_17_08_899Z.png} & \includegraphics[trim={ 0 48 0 384},clip,width=0.32\linewidth]{interpolbefore.png} \\[-3.5pt]
      
      \raisebox{12pt}{EAE}
      & \includegraphics[trim={ 0 0 0 640},clip,width=0.32\linewidth]{reconstructionsceleba.png}&\includegraphics[trim={ 0 0 0 576},clip,width=0.32\linewidth]{image_2020_02_18T14_17_08_899Z.png} & \includegraphics[trim={ 0 0 0 432},clip,width=0.32\linewidth]{interpolbefore.png} \\

      \bottomrule

    \end{tabular}
\end{sc}}
  \vspace{2.5mm}
  \caption{Qualitative comparisons to RAE variants and other standard benchmarks on CelebA. On the left, we have reconstructions (top row being ground truth GT) , the middle has generated samples, the right has interpolations. From top to bottom ignoring GT: VAE, CV-VAE, WAE, 2SVAE, RAE-GP, RAE-L2, RAE-SN, RAE, AE, EAE. Non-EAE figures reproduced from \citep{ghosh2019variational} }
  \label{RAEcomparison}
\end{figure*}

While Table \ref{FIDscores} captures the quantitative performance of our method, we seek to provide a qualitative comparison as well. This is done in Figure \ref{RAEcomparison}. We compare to all RAE variants, as well as 2SVAE, WAE, CV-VAE and the standard VAE and AE as in Table \ref{FIDscores}. Results for CIFAR-10 and MNIST appear in the supplementary material.

\subsection{GAUSSIAN LATENTS WITHOUT ENTROPIC REGULARIZATION}

\begin{figure*}[htp]

\centering
\includegraphics[width=.3\textwidth]{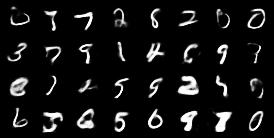}\hfill
\includegraphics[width=.3\textwidth]{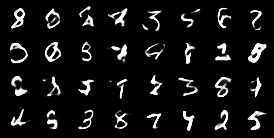}\hfill
\includegraphics[width=.3\textwidth]{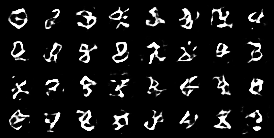}

\caption{Variation in bottleneck width causes massive differences in generative quality without regularization. From left to right, we present samples ( from $\mathcal{N}(0,I)$ for $8,16,32$ dimensional latent spaces ) }
\label{MNISTwidthvariation}

\end{figure*}

A surprising result emerges as we make the latent space dimensionality lower while ensuring a complex enough decoder and encoder. Though we discussed this process earlier in the context of depth, our architectures are convolutional and a better heuristic proxy is the number of channels while keeping the depth constant. We note that all our encoders share a power of $2$ framework, i.e. channels double every layer from $128$. Keeping this doubling structure, we investigate the effect of width on the latent space with no entropic regularizer. We set the channels to double from $64$, i.e. $64,128,256,512$ and correspondingly in the decoder for MNIST. Figure \ref{MNISTwidthvariation} shows the samples with the latent dimension being set to $8$, and the result when we take corresponding samples from an isotropic Gaussian when the number of latents is $32$.

There is a large, visually evident drop in sample quality by going from a narrow autoencoder to a wide one for generation, when no constraints on the latent space are employed. To confirm the analysis, we provide the result for $16$ dimensions in the figure as well, which is intermediate in quality.

\begin{figure}[h]
    \centering
    \includegraphics[keepaspectratio,width=5 cm]{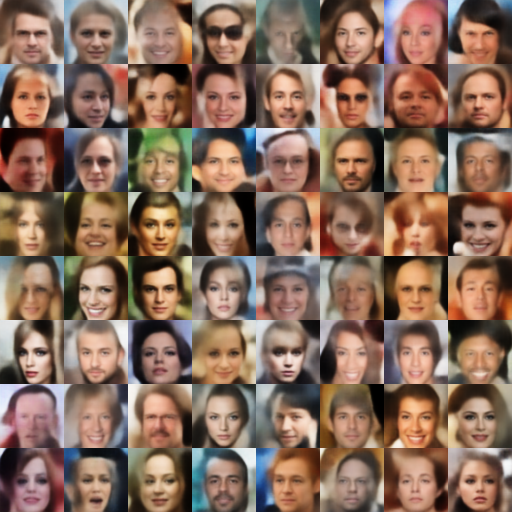}
    \caption{Generated images on CelebA with a narrow bottleneck of $48$, unregularized. The associated FID score was $53.82$.}
    \label{Decent CelebA}
\end{figure}

\begin{figure}[h]
    \centering
    \includegraphics[keepaspectratio,width=5 cm]{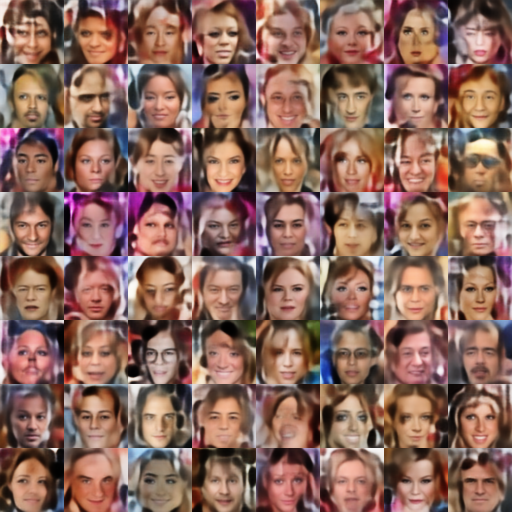}
    \caption{Generated images on CelebA with latent dimensions of $128$, also unregularized. This associates a FID score of $64.72$.}
    \label{Horrible CelebA}
\end{figure}

The aforesaid effect is not restricted to MNIST. We perform a similar study on CelebA taking the latent space from $48$ to $128$, and the results in Figures~\ref{Decent CelebA} and~\ref{Horrible CelebA} show a corresponding change in sample quality. Of course, the results with $48$ dimensional unregularized latents are worse than our regularized, $64$ dimensional sample collage in Figure \ref{CelebA}, but they retain facial quality without artifacts. FID scores (provided in caption) also follow this trend.

In our formulation of the MaxEnt principle, we considered more complex maps (e.g., deeper or wider networks with possibly more channels) able to induce more arbitrary deformations between a latent space and the target space. A narrower bottleneck incentivizes Gaussianization - with a stronger bottleneck, each latent carries more information, with higher entropy in codes $Z$, as discussed in our parallels with Information Bottleneck-like methods. 

\begin{figure}[h]
    \centering
    \includegraphics[keepaspectratio,width=5.5 cm]{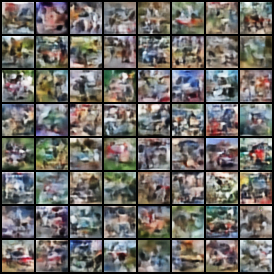}
    \caption{Generated images on CIFAR-10 with unregularized latent dimension of $128$. The FID score is $100.62$, with L2 regularization.}
    \label{Bad cifar}
\end{figure}

\begin{figure}[h]
    \centering
    \includegraphics[keepaspectratio,width=5.5 cm]{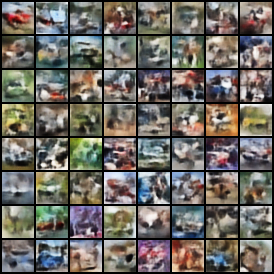}
    \caption{Generated images on CIFAR-10 with unregularized latent dimension equal to $64$. The FID score associated with this checkpoint is $87.45$, trained using L2 regularization.}
    \label{Good cifar}
\end{figure}

We present a similar analysis between CIFAR-10 AEs without regularization. Unlike previous cases, CIFAR-10 samples suffer from the issue that visual quality is less evident to the human eye. These figures are presented in Figures \ref{Bad cifar} and \ref{Good cifar}. The approximate FID difference between these two images is roughly $13$ points ($\approx 100$ vs $\approx 87$). While FID scores are not meaningful for MNIST, we can compare CelebA and CIFAR-10 in terms of FID scores (provided in figure captions). These back up our assertions. For all comparisons, only the latent space is changed and the best checkpoint is taken for both models - we have a case of a less complex model outperforming another that can't be due to more channels allowing for better reconstruction, explainable in MaxEnt terms. 

\section{CONCLUSIONS AND FUTURE WORK}
\label{sec:conclusion}

The VAE has remained a popular deep generative model, while drawing criticism for its blurry images, posterior collapse and other issues. Deterministic encoders have been posited to escape blurriness, since they `lock' codes into a single choice for each instance. We consider our work as reinforcing Wasserstein autoencoders and other recent work in deterministic autoencoders such as RAEs \citep{ghosh2019variational}. In particular, we consider our method of raising entropy to be generalizable whenever batch normalization exists, and note that it solves a more specific problem than reducing the KL between two arbitrary distributions $P,Q$, examining only the case where $P,Q$ satisfy moment constraints. Such reductions can make difficult problems tractable via simple estimators.

We had initially hoped to obtain results via sampling from a prior distribution that were, without ex-post density estimation, already state of the art. In practice, we observed that using a GMM to fit the density improves results, regardless of architecture. These findings might be explained in light of the 2-stage VAE analysis \citep{dai2019diagnosing}, wherein it is postulated that single-stage VAEs inherently struggle to capture Gaussian latents, and a second stage is amenable. To this end, we might aim to design a 2-stage EAE. Numerically, we found such an architecture hard to tune, as opposed to a single stage EAE which was robust to the choice of hyperparameters. We believe this might be an interesting future direction.

We note that our results improve on the RAE, which in turn improved on the 2SVAE FID numbers. Though the latest GAN architectures remain out of reach in terms of FID scores for most VAE models, 2SVAE came within striking distance of older ones, such as the vanilla WGAN. Integrating state of the art techniques for VAEs as in, for instance, VQVAE2 \citep{razavi2019generating} to challenge GAN-level benchmarks could form an interesting future direction. Quantized latent spaces also offer a more tractable framework for entropy based models and allow us to work with discrete entropy which is a more meaningful function. As noted earlier, we do not compare to the ResNet equipped bigWAE models \citep{tolstikhin2017wasserstein}, which are far larger but also deliver better results (up to $35$ for CelebA).

The previous work on RAEs \citep{ghosh2019variational}, which our method directly draws on deserves special addressal. The RAE method shows that deterministic autoencoders can succeed at generation, so long as regularizers are applied and post-density estimation is carried out. Yet, while regularization is certainly nothing out of the ordinary, the density estimation step robs RAEs of sampling from any isotropic prior. We improve on the RAE techniques when density estimation is in play, but more pertinently, we keep a method for isotropic sampling that is rigorously equivalent to cross entropy minimization. As such, we offer better performance while adding more features, and our isotropic results far outperform comparable isotropic benchmarks.

\section*{Acknowledgments}

Resources used in preparing this research were provided by NSERC, the Province of Ontario, the Government of Canada through CIFAR, and companies sponsoring the Vector Institute\footnote{www.vectorinstitute.ai/\#partners}. We thank Yaoliang Yu for discussions.

\bibliography{example_paper}
\bibliographystyle{plainnat}

\clearpage

\section*{Appendix containing supplementary material and additional results}

On the next two pages, we present additional qualitative results.

We add in this page general notes on the training of EAEs, as requested by reviewers. In particular, across extensive experiments, we noted the following empirical trends and heuristics which we choose to pass on for the sake of ease of implementation.

\begin{itemize}
    \item Results for all EAEs that use the Kozachenko-Leonenko or similar KNN based entropy estimators can be improved in general by using $\geq 2$ neighbours per minibatch. However, we do not recommend this. Performance gains from this are slight, and for most applications, using $1$ suffices.
    \item We recommend using at least $3$ conv-deconv layers in the encoder and decoder for any autoencoding pair for all three datasets for best FID scores.
    \item For both CIFAR and CelebA, we recommend a minimum latent size of $32$.
    \item For CIFAR-10 in particular, learning rate decay is critical when using the ADAM optimizer. We use an exponential decay with a decay rate $\geq 0.98$. It should be noted that \citep{ghosh2019variational} use a more complex schedule that involves looking at the validation loss. We did not require such.
    \item Good samples emerge early - samples for all three datasets generated by epoch $10$ as evaluated by a human eye are highly predictive of eventual best performance in terms of FID. As such, it is recommended to periodically generate samples and visually inspect them.
\end{itemize}

\subsection*{Preprocessing datasets}

Here, we detail the pre-processing of datasets common to our methods and the methods we benchmark against. We carry out no pre-processing for CIFAR-10. For MNIST, we pad with zeros to reach $32 \times 32$ as the shape. For CelebA, pre-processing is important and can vastly change FID scores. We perform a center-crop to $140 \times 140$ before resizing to $64 \times 64$.

\subsection*{Details of following material}

In Table 1 below, we present a larger version of the FID results from Table \ref{FIDscores} in the main paper.  In Figures \ref{CIFAR comparison} and \ref{MNISTcomparison} below, we also present qualitative results including reconstruction and interpolations on the latent space that serve to show that the latent spaces obtained by EAEs are meaningful. These experiments on latent spaces mirror \citep{ghosh2019variational}.

\begin{table*}
\centering
\begin{tabular}{l|ll|ll}
\hline
& CIFAR-10 & & CelebA & \\
\hline
Architectures(Isotropic)  & FID  & Reconstruction & FID & Reconstruction  \\
\hline
VAE       & 106.37  & 57.94  & 48.12 &  39.12    \\
CV-VAE       &  94.75  & 37.74   & 48.87 &  40.41  \\
WAE    &  117.44  &  35.97  & 53.67& \textbf{34.81}     \\
2SVAE   &  109.77  & 62.54  & 49.70 &  42.04  \\
EAE & {\bf 85.26(84.53)} & 29.77  & {\bf 44.63} &40.26   \\
\hline
Architectures(MVG)  & FID  & Reconstruction & FID & Reconstruction \\
\hline
RAE   &  83.87  & 29.05   & 48.20 & 40.18     \\
RAE-L2  &  80.80  & 32.24    & 51.13 &  43.52    \\
RAE-GP     &  83.05 & 32.17  &  116.30  &   39.71    \\
RAE-SN    &   84.25 & \textbf{27.61}  & 44.74  &   36.01   \\
AE     &  84.74 & 30.52 & 127.85   &   40.79    \\
AE-L2  &  247.48 & 34.35  & 346.29  &   44.72   \\

EAE &  {\bf 80.07} & 29.77  & {\bf 42.92} &  40.26  \\
\hline
Architectures(GMM)  & FID  & Reconstruction & FID & Reconstruction \\
\hline
VAE   &  103.78  & 57.94   & 45.52 & 39.12     \\
CV-VAE   &  86.64  & 37.74   & 49.30 & 40.41     \\
WAE  &  93.53  & 35.97   & 42.73 & \textbf{34.81}    \\
2SVAE   &  N/A  & 62.54   & N/A & 42.04     \\
RAE   &  76.28  & 29.05   & 44.68 & 40.18      \\
RAE-L2  &  74.16  & 32.24    & 47.97 &  43.52    \\
RAE-GP     &  76.33 & 32.17  &  45.63  &   39.71    \\
RAE-SN    &   75.30 &  \textbf{27.61}  & 40.95  &  36.01   \\
AE     &  76.47 & 30.52 & 45.10   &   40.79    \\
AE-L2  &  75.40 & 34.35  & 48.42  &   44.72   \\

EAE &  {\bf 73.12} & 29.77  & {\bf 39.76} &  40.26  \\
\hline
\end{tabular}
\caption{FID scores for relevant VAEs and VAE-like architectures. Scores within parentheses for EAE denote a regularization on a linear map. Isotropic denotes samples drawn from a latent space of $\mathcal{N} (0,I)$. GMM denotes sampling from a mixture of $10$ Gaussians of full covariance. These evaluations correspond to analogous benchmarking for RAEs \citep{ghosh2019variational}.  Alongside FID values appearing in Table 1 of the main paper, we add results obtained when a Multivariate Gaussian (MVG) i.e. $\mathcal{N} (\mu,\Sigma)$ of full covariance is used for ex-post density estimation. Note that values for reconstruction are not changed by change of density estimators.}
\label{FIDscoressupp}
\end{table*}

\begin{figure*}[htp]
  \centering
  \scalebox{.9}
{\setlength\tabcolsep{3pt}
\begin{sc}
  \small
    \begin{tabular}{ r c c c}
    \toprule
        & Reconstructions & Random Samples & Interpolations \\
        \midrule
      \raisebox{10pt}{GT}
      & \includegraphics[trim={ 0 320 0
        0},clip,width=0.32\linewidth]{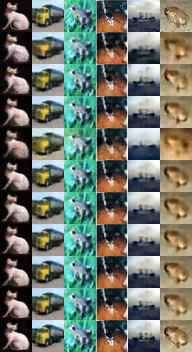}
      \\[-3.5pt]

      \raisebox{12pt}{VAE}
      & \includegraphics[trim={ 0 288 0 32},clip,width=0.32\linewidth]{cifarrecon.png}&\includegraphics[trim={0 216 0 24},clip,width=0.32\linewidth]{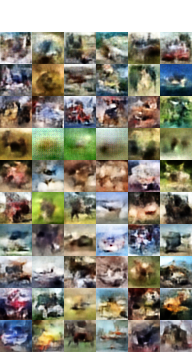} & \includegraphics[trim={ 0 216 0 0},clip,width=0.32\linewidth]{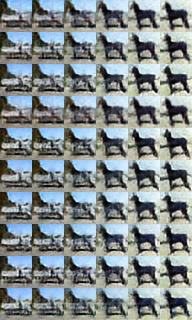} \\[-3.5pt]
      
            \raisebox{12pt}{CV-VAE}
      & \includegraphics[trim={ 0 256 0 64},clip,width=0.32\linewidth]{cifarrecon.png}&\includegraphics[trim={ 0 192 0 48},clip,width=0.32\linewidth]{cifarsample1.png} & \includegraphics[trim={ 0 192 0 24},clip,width=0.32\linewidth]{cifarinterpol.png} \\[-3.5pt]
      
            \raisebox{12pt}{WAE}
      & \includegraphics[trim={ 0 224 0 96},clip,width=0.32\linewidth]{cifarrecon.png}&\includegraphics[trim={ 0 168 0 72},clip,width=0.32\linewidth]{cifarsample1.png} & \includegraphics[trim={ 0 168 0 48},clip,width=0.32\linewidth]{cifarinterpol.png} \\[-3.5pt]
      
            \raisebox{12pt}{2SVAE}
      & \includegraphics[trim={ 0 192 0 128},clip,width=0.32\linewidth]{cifarrecon.png}&\includegraphics[trim={ 0 144 0 96},clip,width=0.32\linewidth]{cifarsample1.png} & \includegraphics[trim={ 0 144 0 72},clip,width=0.32\linewidth]{cifarinterpol.png} \\[-3.5pt]
      
            \raisebox{12pt}{RAE-GP}
      & \includegraphics[trim={ 0 160 0 160},clip,width=0.32\linewidth]{cifarrecon.png}&\includegraphics[trim={ 0 120 0 120},clip,width=0.32\linewidth]{cifarsample1.png} & \includegraphics[trim={ 0 120 0 96},clip,width=0.32\linewidth]{cifarinterpol.png} \\[-3.5pt]
      
            \raisebox{12pt}{RAE-L2}
      & \includegraphics[trim={ 0 128 0 192},clip,width=0.32\linewidth]{cifarrecon.png}&\includegraphics[trim={ 0 96 0 144},clip,width=0.32\linewidth]{cifarsample1.png} & \includegraphics[trim={ 0 96 0 120},clip,width=0.32\linewidth]{cifarinterpol.png} \\[-3.5pt]
      
            \raisebox{12pt}{RAE-SN}
      & \includegraphics[trim={ 0 96 0 224},clip,width=0.32\linewidth]{cifarrecon.png}&\includegraphics[trim={ 0 72 0 168},clip,width=0.32\linewidth]{cifarsample1.png} & \includegraphics[trim={ 0 72 0 144},clip,width=0.32\linewidth]{cifarinterpol.png} \\[-3.5pt]
      
            \raisebox{12pt}{RAE}
      & \includegraphics[trim={ 0 64 0 256},clip,width=0.32\linewidth]{cifarrecon.png}&\includegraphics[trim={ 0 48 0 192},clip,width=0.32\linewidth]{cifarsample1.png} & \includegraphics[trim={ 0 48 0 168},clip,width=0.32\linewidth]{cifarinterpol.png} \\[-3.5pt]
      
            \raisebox{12pt}{AE}
      & \includegraphics[trim={ 0 32  0 288},clip,width=0.32\linewidth]{cifarrecon.png}&\includegraphics[trim={ 0 24  0 216},clip,width=0.32\linewidth]{cifarsample1.png} & \includegraphics[trim={ 0 24  0 192},clip,width=0.32\linewidth]{cifarinterpol.png} \\[-3.5pt]
      
            \raisebox{12pt}{EAE}
      & \includegraphics[trim={ 0 0 0 320},clip,width=0.32\linewidth]{cifarrecon.png}&\includegraphics[trim={ 0 0 0 240},clip,width=0.32\linewidth]{cifarsample1.png} & \includegraphics[trim={ 0 0 0 216},clip,width=0.32\linewidth]{cifarinterpol.png} \\

      \raisebox{10pt}{GT}
      & \includegraphics[trim={ 0 240 0
        0},clip,width=0.32\linewidth]{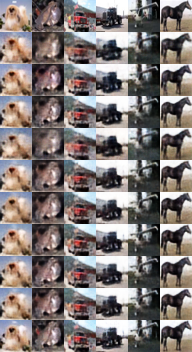}
      \\[-3.5pt]

      \raisebox{12pt}{VAE}
      & \includegraphics[trim={ 0 216 0 24},clip,width=0.32\linewidth]{cifarrecon2.png}&\includegraphics[trim={0 216 0 24},clip,width=0.32\linewidth]{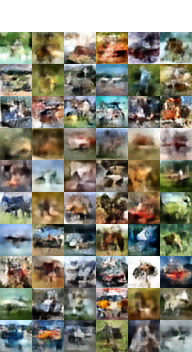} & \includegraphics[trim={ 0 216 0 0},clip,width=0.32\linewidth]{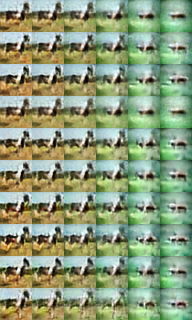} \\[-3.5pt]
      
            \raisebox{12pt}{CV-VAE}
      & \includegraphics[trim={ 0 192 0 48},clip,width=0.32\linewidth]{cifarrecon2.png}&\includegraphics[trim={ 0 192 0 48},clip,width=0.32\linewidth]{cifarsample2.png} & \includegraphics[trim={ 0 192 0 24},clip,width=0.32\linewidth]{cifarinterpol2.png} \\[-3.5pt]
      
            \raisebox{12pt}{WAE}
      & \includegraphics[trim={ 0 168 0 72},clip,width=0.32\linewidth]{cifarrecon2.png}&\includegraphics[trim={ 0 168 0 72},clip,width=0.32\linewidth]{cifarsample2.png} & \includegraphics[trim={ 0 168 0 48},clip,width=0.32\linewidth]{cifarinterpol2.png} \\[-3.5pt]
      
            \raisebox{12pt}{2SVAE}
      & \includegraphics[trim={ 0 144 0 96},clip,width=0.32\linewidth]{cifarrecon2.png}&\includegraphics[trim={ 0 144 0 96},clip,width=0.32\linewidth]{cifarsample2.png} & \includegraphics[trim={ 0 144 0 72},clip,width=0.32\linewidth]{cifarinterpol2.png} \\[-3.5pt]
      
            \raisebox{12pt}{RAE-GP}
      & \includegraphics[trim={ 0 120 0 120},clip,width=0.32\linewidth]{cifarrecon2.png}&\includegraphics[trim={ 0 120 0 120},clip,width=0.32\linewidth]{cifarsample2.png} & \includegraphics[trim={ 0 120 0 96},clip,width=0.32\linewidth]{cifarinterpol2.png} \\[-3.5pt]
      
            \raisebox{12pt}{RAE-L2}
      & \includegraphics[trim={ 0 96 0 144},clip,width=0.32\linewidth]{cifarrecon2.png}&\includegraphics[trim={ 0 96 0 144},clip,width=0.32\linewidth]{cifarsample2.png} & \includegraphics[trim={ 0 96 0 120},clip,width=0.32\linewidth]{cifarinterpol2.png} \\[-3.5pt]
      
            \raisebox{12pt}{RAE-SN}
      & \includegraphics[trim={ 0 72 0 168},clip,width=0.32\linewidth]{cifarrecon2.png}&\includegraphics[trim={ 0 72 0 168},clip,width=0.32\linewidth]{cifarsample2.png} & \includegraphics[trim={ 0 72 0 144},clip,width=0.32\linewidth]{cifarinterpol2.png} \\[-3.5pt]
      
            \raisebox{12pt}{RAE}
      & \includegraphics[trim={ 0 48 0 192},clip,width=0.32\linewidth]{cifarrecon2.png}&\includegraphics[trim={ 0 48 0 192},clip,width=0.32\linewidth]{cifarsample2.png} & \includegraphics[trim={ 0 48 0 168},clip,width=0.32\linewidth]{cifarinterpol2.png} \\[-3.5pt]
      
            \raisebox{12pt}{AE}
      & \includegraphics[trim={ 0 24  0 216},clip,width=0.32\linewidth]{cifarrecon2.png}&\includegraphics[trim={ 0 24  0 216},clip,width=0.32\linewidth]{cifarsample2.png} & \includegraphics[trim={ 0 24  0 192},clip,width=0.32\linewidth]{cifarinterpol2.png} \\[-3.5pt]
      
            \raisebox{12pt}{EAE}
      & \includegraphics[trim={ 0 0 0 240},clip,width=0.32\linewidth]{cifarrecon2.png}&\includegraphics[trim={ 0 0 0 240},clip,width=0.32\linewidth]{cifarsample2.png} & \includegraphics[trim={ 0 0 0 216},clip,width=0.32\linewidth]{cifarinterpol2.png} \\

      \bottomrule

    \end{tabular}
\end{sc}}
  \vspace{2.5mm}
  \caption{Qualitative comparisons to RAE variants and other standard benchmarks on CIFAR-10. On the left, we have reconstructions (top row being ground truth GT), the middle has generated samples, the right has interpolations. From top to bottom ignoring GT: VAE, CV-VAE, WAE, 2SVAE, RAE-GP, RAE-L2, RAE-SN, RAE, AE, EAE. Non-EAE figures reproduced from \citep{ghosh2019variational} }
  \label{CIFAR comparison}
\end{figure*}

\begin{figure*}[h]
  \centering
  \scalebox{.9}
{\setlength\tabcolsep{3pt}
\begin{sc}
  \small
    \begin{tabular}{ r c c c}
    \toprule
        & Reconstructions & Random Samples & Interpolations \\
        \midrule
      \raisebox{10pt}{GT}
      & \includegraphics[trim={ 0 240 0
        0},clip,width=0.32\linewidth]{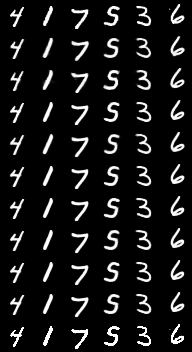}
      \\[-3.5pt]

      \raisebox{12pt}{VAE}
      & \includegraphics[trim={ 0 216 0 24},clip,width=0.32\linewidth]{mnistrecon.png}&\includegraphics[trim={0 216 0 24},clip,width=0.32\linewidth]{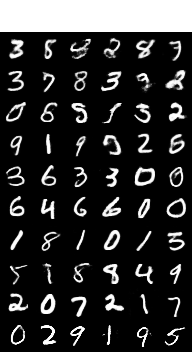} & \includegraphics[trim={ 0 216 0 24},clip,width=0.32\linewidth]{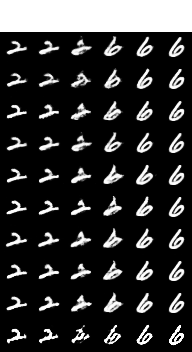} \\[-3.5pt]
      
            \raisebox{12pt}{CV-VAE}
      & \includegraphics[trim={ 0 192 0 48},clip,width=0.32\linewidth]{mnistrecon.png}&\includegraphics[trim={ 0 192 0 48},clip,width=0.32\linewidth]{mnistsamples.png} & \includegraphics[trim={ 0 192 0 48},clip,width=0.32\linewidth]{mnistinter.png} \\[-3.5pt]
      
            \raisebox{12pt}{WAE}
      & \includegraphics[trim={ 0 168 0 72},clip,width=0.32\linewidth]{mnistrecon.png}&\includegraphics[trim={ 0 168 0 72},clip,width=0.32\linewidth]{mnistsamples.png} & \includegraphics[trim={ 0 168 0 72},clip,width=0.32\linewidth]{mnistinter.png} \\[-3.5pt]
      
            \raisebox{12pt}{2SVAE}
      & \includegraphics[trim={ 0 144 0 96},clip,width=0.32\linewidth]{mnistrecon.png}&\includegraphics[trim={ 0 144 0 96},clip,width=0.32\linewidth]{mnistsamples.png} & \includegraphics[trim={ 0 144 0 96},clip,width=0.32\linewidth]{mnistinter.png} \\[-3.5pt]
      
            \raisebox{12pt}{RAE-GP}
      & \includegraphics[trim={ 0 120 0 120},clip,width=0.32\linewidth]{mnistrecon.png}&\includegraphics[trim={ 0 120 0 120},clip,width=0.32\linewidth]{mnistsamples.png} & \includegraphics[trim={ 0 120 0 120},clip,width=0.32\linewidth]{mnistinter.png} \\[-3.5pt]
      
            \raisebox{12pt}{RAE-L2}
      & \includegraphics[trim={ 0 96 0 144},clip,width=0.32\linewidth]{mnistrecon.png}&\includegraphics[trim={ 0 96 0 144},clip,width=0.32\linewidth]{mnistsamples.png} & \includegraphics[trim={ 0 96 0 144},clip,width=0.32\linewidth]{mnistinter.png} \\[-3.5pt]
      
            \raisebox{12pt}{RAE-SN}
      & \includegraphics[trim={ 0 72 0 168},clip,width=0.32\linewidth]{mnistrecon.png}&\includegraphics[trim={ 0 72 0 168},clip,width=0.32\linewidth]{mnistsamples.png} & \includegraphics[trim={ 0 72 0 168},clip,width=0.32\linewidth]{mnistinter.png} \\[-3.5pt]
      
            \raisebox{12pt}{RAE}
      & \includegraphics[trim={ 0 48 0 192},clip,width=0.32\linewidth]{mnistrecon.png}&\includegraphics[trim={ 0 48 0 192},clip,width=0.32\linewidth]{mnistsamples.png} & \includegraphics[trim={ 0 48 0 192},clip,width=0.32\linewidth]{mnistinter.png} \\[-3.5pt]
      
            \raisebox{12pt}{AE}
      & \includegraphics[trim={ 0 24  0 216},clip,width=0.32\linewidth]{mnistrecon.png}&\includegraphics[trim={ 0 24  0 216},clip,width=0.32\linewidth]{mnistsamples.png} & \includegraphics[trim={ 0 24  0 216},clip,width=0.32\linewidth]{mnistinter.png} \\[-3.5pt]
      
            \raisebox{12pt}{EAE}
      & \includegraphics[trim={ 0 0 0 240},clip,width=0.32\linewidth]{mnistrecon.png}&\includegraphics[trim={ 0 0 0 240},clip,width=0.32\linewidth]{mnistsamples.png} & \includegraphics[trim={ 0 0 0 240},clip,width=0.32\linewidth]{mnistinter.png} \\

      \bottomrule

    \end{tabular}
\end{sc}}
  \vspace{2.5mm}
  \caption{Qualitative comparisons to RAE variants and other standard benchmarks on MNIST. On the left, we have reconstructions (top row being ground truth GT) , the middle has generated samples, the right has interpolations. From top to bottom ignoring GT: VAE, CV-VAE, WAE, 2SVAE, RAE-GP, RAE-L2, RAE-SN, RAE, AE, EAE. Non-EAE figures reproduced from \citep{ghosh2019variational}}
  \label{MNISTcomparison}
\end{figure*}

\clearpage

\end{document}